\documentclass{article}

     \PassOptionsToPackage{round,authoryear}{natbib}

 \usepackage[preprint]{neurips_2023}

\usepackage[utf8]{inputenc} 
\usepackage[T1]{fontenc}    
\usepackage{hyperref}       
\usepackage{url}            
\usepackage{booktabs}       
\usepackage{amsfonts}       
\usepackage{nicefrac}       
\usepackage{microtype}      
\usepackage{xcolor}         
\usepackage{bbm}
\usepackage{amsmath}
\usepackage{multirow}
\usepackage[normalem]{ulem}
\usepackage{graphics}
\useunder{\uline}{\ul}{}
\usepackage{graphicx}
\usepackage{wrapfig}
\usepackage{amssymb}
\usepackage{tabularx}
\usepackage{fancyvrb}
\usepackage{listings}

\usepackage{scrextend}
\usepackage{xspace}

\def\modelname{\textsc{Law}\xspace}

\title{Language Models, Agent Models, and World Models:\\
The LAW for Machine Reasoning and Planning}

\author{
Zhiting Hu\thanks{Equal contribution}~~ (UCSD),~~~ Tianmin Shu$^*$ (JHU) \\
\texttt{zhh019@ucsd.edu,~~~ tianmin.shu@jhu.edu}
}

\begin{document}

\maketitle

\begin{abstract}
Despite their tremendous success in many applications, large language models often fall short of consistent reasoning and planning in various (language, embodied, and social) scenarios, due to inherent limitations in their inference, learning, and modeling capabilities. In this position paper, we present a new perspective of machine reasoning, \modelname, that connects the concepts of \uline{l}anguage models, \uline{a}gent models, and \uline{w}orld models, for more robust and versatile reasoning capabilities. In particular, we propose that world and agent models are a better {\it abstraction} of reasoning, that introduces the crucial elements of deliberate human-like reasoning, including beliefs about the world and other agents, anticipation of consequences, goals/rewards, and strategic planning. Crucially, language models in \modelname serve as a {\it backend} to implement the system or its elements and hence provide the computational power and adaptability. We review the recent studies that have made relevant progress and discuss future research directions towards operationalizing the \modelname framework.
\end{abstract}

\section{Introduction}\label{sec:intro}

Large language models (LLMs) are among the most powerful intelligent machines people have built to date. They are adept at generating natural language continuations from a given text (or multi-modal) input. Natural language is a flexible means for humans to describe the world, express thoughts, and communicate with each other. LLMs, trained with the vast text humans have ever produced, inherit much of the knowledge conveyed through natural language, including the causal structure of the world (expressed in phrases like {\it ``a bottle is pushed, water pours out''}), reasonings about various subjects, scientific theories, beliefs, cultural norms, etc. 

\begin{figure}
    \centering
    \includegraphics[width=\linewidth]{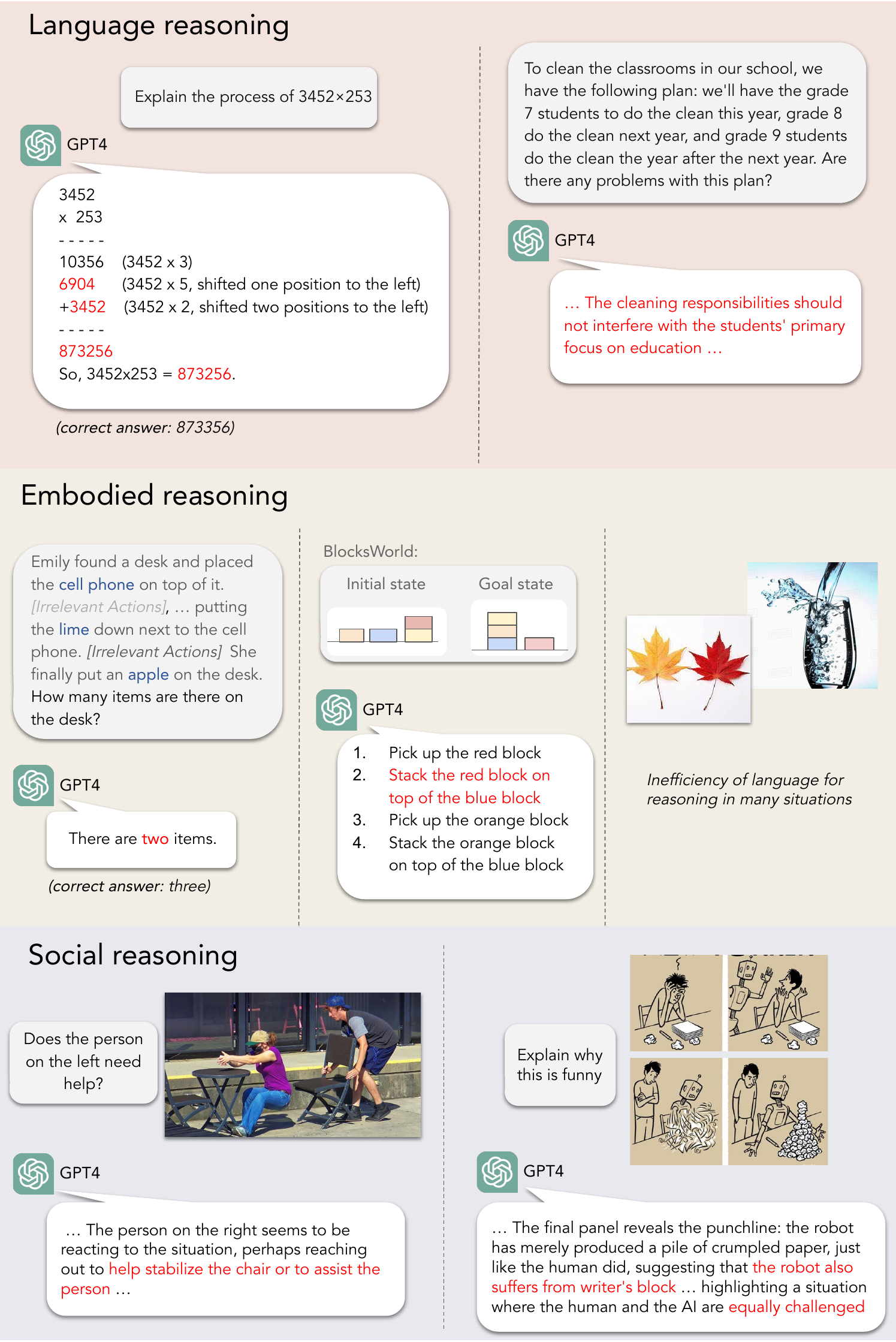}
    \caption{LLMs, such as GPT4, fail in simple tasks of language, embodied, and social reasoning. Erroneous parts in the answers are highlighted in {\color{red} red}.}
    \label{fig:failures}
\end{figure}

On the other hand, LLMs often fall short of consistent reasoning and planning and sometimes fail surprisingly in tasks that humans find easy. Figure~\ref{fig:failures} shows such examples in different reasoning scenarios. These failure examples highlight several fundamental limitations of machine reasoning based on LLMs:

First, natural language text is often ambiguous and imprecise. One of the key reasons for this ambiguity and imprecision is that the rich context, which humans rely on when producing the text, is often missing. This context includes the specific perceptual and social situations the human agents were in, their mental states (e.g., intentions,  beliefs, and thinking processes), and world commonsense. Thus LLMs, which learn only to simulate the surface text without modeling the underlying context, lack {\it grounding} on the physical, social, and mental experiences. 

Another core limitation of LLMs arises from the {\it inefficiency of language} as the medium for carrying out reasoning in certain situations (Figure~\ref{fig:failures}, embodied reasoning). For instance, articulating all subtle differences between two leaves might require an extensive text paragraph. In contrast, generating an image that visually represents these leaves can be far more efficient, requiring just a few pixels. Similarly, using other sensory modalities (e.g., videos) is often more straightforward than relying on language to describe intuitive physics, such as predicting fluid flow based on its viscosity and the surrounding obstacles. 





These limitations are further exacerbated by the inference procedures of LLMs. They reason by generating text autoregressively, token-by-token, from left to right in a single pass, reminiscent of humans' System-I intuitive thinking. Humans' System-II reasoning stands in stark contrast to LLM reasoning. 
In particular, humans possess a mental model of the world. The ``{\it world model}'' in our minds enables us to simulate actions and their effects on the world's state for robust reasoning during complex tasks \citep{tolman1948cognitive, briscoe2011mental, battaglia2013simulation, allen2020rapid, pramod2020evidence}. For example, when planning to achieve a goal, we use our internal world model to think about different actions we could take and predict possible outcomes for each choice. This prediction of outcomes in turn helps refine the action plan for better attaining the goal. This decision-making process is governed by an ``{\it agent model}'' on top of the world model. Further, in social reasoning tasks, human agents additionally use their {\it beliefs} about other agents. For example, during a conversation, an agent needs to infer others' intentions and their potential reactions to decide the most appropriate things to say. 
Therefore, humans achieve their goals and successfully interact with one another through deliberate planning guided by their internal models of the world and other agents.


%

\begin{figure}[t]
    \centering
    \includegraphics[width=\linewidth]{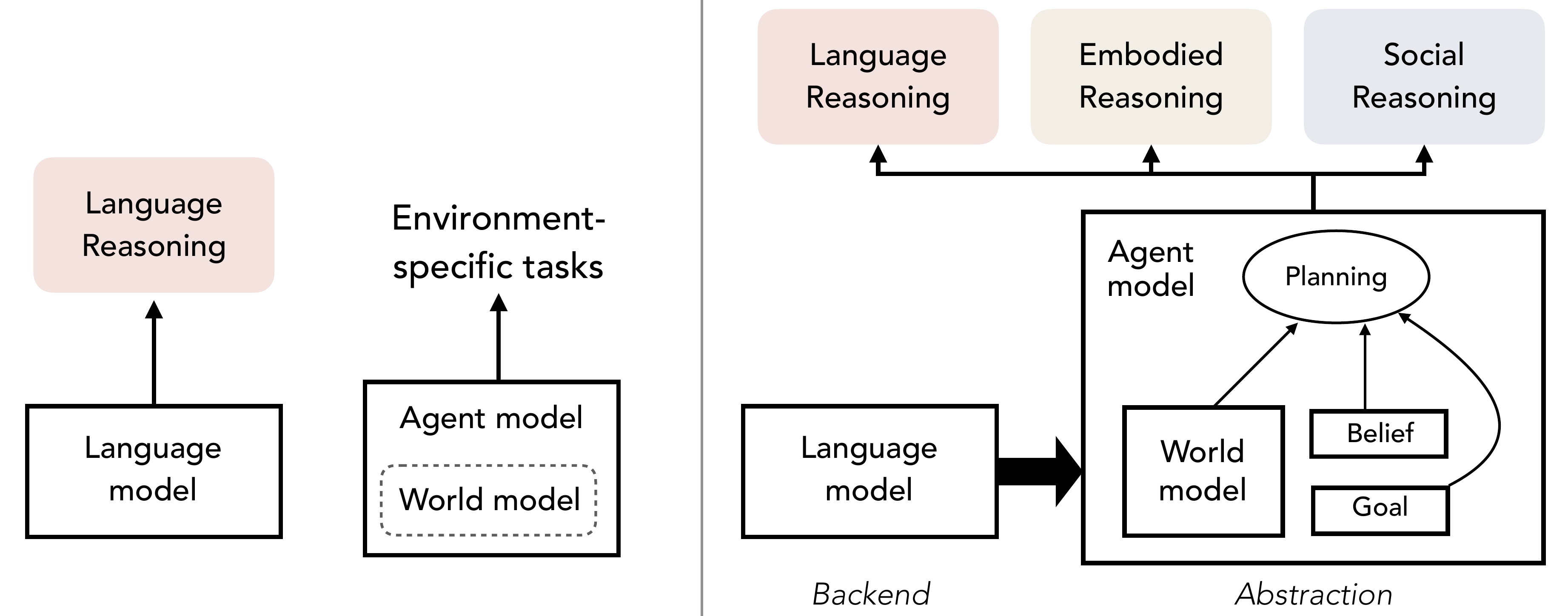}
    \caption{{\bf Left:} Language models and world/agent models are usually studied in different contexts. {\bf Right:} The proposed \modelname framework for more general and robust reasoning, with world and agent models as the abstraction of reasoning and language models as the backend implementation.}
    \label{fig:law}
    \vspace{-1pt}
\end{figure}

Human agents also exhibit richer learning mechanisms than LLMs. As shown in Figure~\ref{fig:failures} (embodied/social reasoning), LLMs trained merely with large-scale text corpora lack fundamental real-world experience, such as tracking and interacting with objects, understanding real-world physics and spatiotemporal relationships, sensing and tracking the world states, and recognizing other agents' behaviors, etc. Human agents bypass these limitations by learning through interaction with the environment. For instance, we acquire new knowledge by attempting tasks and receiving feedback (e.g., a chef refines their culinary skills by experimenting with different ingredients and tasting the outcomes), or simply by exploring the surroundings randomly (e.g., a child learns about different textures and sensations by randomly picking up various objects). 

In sum, current LLM reasoning and planning face key limitations in {inference} (autoregressive generation), {learning} (imitation from data corpora without real-world interaction), and {modeling} (inefficiency of language and its lack of grounding). In this position paper, we present a new perspective toward more general and robust machine reasoning across language, embodied, social, and other broad scenarios. In particular, inspired by the above discussion, we propose a unified \modelname framework for machine reasoning that connects the concepts of \uline{l}anguage models, \uline{a}gent models, and \uline{w}orld models (Figure~\ref{fig:law}, right). 


Specifically, the concepts of world and agent models have their roots in cognitive science and developmental psychology \citep[e.g.,][]{tolman1948cognitive,premack1978does,johnson1983mental,johnson2010mental, gentner2014mental,nortmann2015primary,maus2013motion,forrester1971counterintuitive,gopnik1994theory, gergely2003teleological,spelke2007core,battaglia2013simulation,baker2009action,jara2016naive,baker2017rational}. As mentioned earlier, a world model (\S\ref{sec:pre:wm}) is a mental representation that agents use to understand and predict the external world around them; an agent model (\S\ref{sec:pre:am}) contains a world model and also other crucial components, including the agent's goals as well as its beliefs of the current world state and other agents. These components together shape the agent's cognitive processes, enabling deliberate reasoning and planning.
In the fields of artificial intelligence and machine learning, world and agent models have typically been studied in the context of reinforcement learning and robotics \citep[e.g.,][]{toussaint2003learning, schulkin2012action, ha2018world, berkenkamp2017safe, clavera2018model, zhang2019solar, kaiser2019model,moerland2023model, lecun2022path}. For instance, recent studies show world modeling enables agents to make effective action plans in specific games and embodied control problems \citep{schrittwieser2020mastering,hafner2020mastering}. 


In this paper, we highlight the enormous new opportunities of integrating language models with world and agent models, for more general reasoning capabilities not possible with the individual formulations alone. In particular, compared to the current paradigm of LM-based reasoning, we posit that world and agent models are a better {\it abstraction} of machine reasoning, as they natively encompass the essential components for human-like reasoning---e.g., beliefs, goals, anticipation of consequences, and deliberate planning (Figure~\ref{fig:law}, right). In this framework, LMs are one of the ways for {\it implementing} world/agent models or the individual components. That is, LMs serve as the backend that operationalizes the framework. Compared to conventional implementations, LMs provide the computational power and adaptability necessary for handling vastly diverse reasoning scenarios. On the other hand, the new role of LMs in the \modelname reasoning framework also highlights their limitations and inspires future research for improvement. 

In the following sections, we first give a brief background of the three models, respectively (\S\ref{sec:pre}). We then present the new \modelname framework of reasoning (\S\ref{sec:law}), where we review the emerging studies related to each element in the framework, and discuss the roadmap for addressing the various challenges inherent in existing approaches and achieving more advanced machine reasoning and planning.

\section{Preliminary: The Three Models}\label{sec:pre}

\subsection{Language Models (LMs)}\label{sec:pre:lm}

A modern neural LM processes text by learning to predict the next word $x_t$ given the preceding text sequence $x_{1:t-1}$:
\begin{equation}
    P(x_t | x_{1:t-1}).
\end{equation}
Pretrained with massive text (and multi-modal) data corpora, LLMs, such as (Chat)GPTs \citep{brown2020language,openai2023gpt4}, Gemini \citep{Gemini2023}, and Llama \citep{touvron2023llama,touvron2023llama2}, have exhibited emergent reasoning abilities in a wide range of language tasks, including question answering, math reasoning, code generation, conversation, and others.

\subsection{World Models (WMs)}\label{sec:pre:wm}


The knowledge of the world is extremely broad, ranging from how a ball would fall and bounce off the ground, to how the price of a stock would rise and fall. In the context of embodied tasks (where the world model concept is usually studied), a world model can typically be  formulated as state transition probabilities, which characterizes a generative, casual mechanism of how the world state changes after an agent's actions:
\begin{equation}
    \mathcal{T}(s^\prime | s, a),
    \label{eq:world_model}
\end{equation}
where $s$ is the current world state, $a$ is an action taken by an agent, and $s^\prime$ is the next state after the action. 

The need for a world model to conduct commonsense physical reasoning \citep{battaglia2013simulation,ullman2017mind,smith2019modeling} and problem-solving such as tool use and model-based planning \citep{allen2020rapid} has long been argued for in cognitive science. There has also been recent evidence from neuroscience suggesting that our brains use a physics engine as a world model to simulate the future \citep{pramod2020evidence}. Similarly, there has been increasing interest in building a world model for physical scene understanding \citep{wu2017learning,li2020visual,allen2022learning} and model-based reinforcement learning \citep{berkenkamp2017safe, clavera2018model, zhang2019solar, kaiser2019model,hafner2020mastering,moerland2023model} and planning \citep{toussaint2018differentiable,li2019learning,jatavallabhula2021gradsim}. These prior works have demonstrated that the use of world models can enable more data-efficient learning and better generalization in unseen scenarios. 


\subsection{Agent Models (AMs)}\label{sec:pre:am}

\begin{figure}[t]
    \centering
    \includegraphics[width=0.6\linewidth]{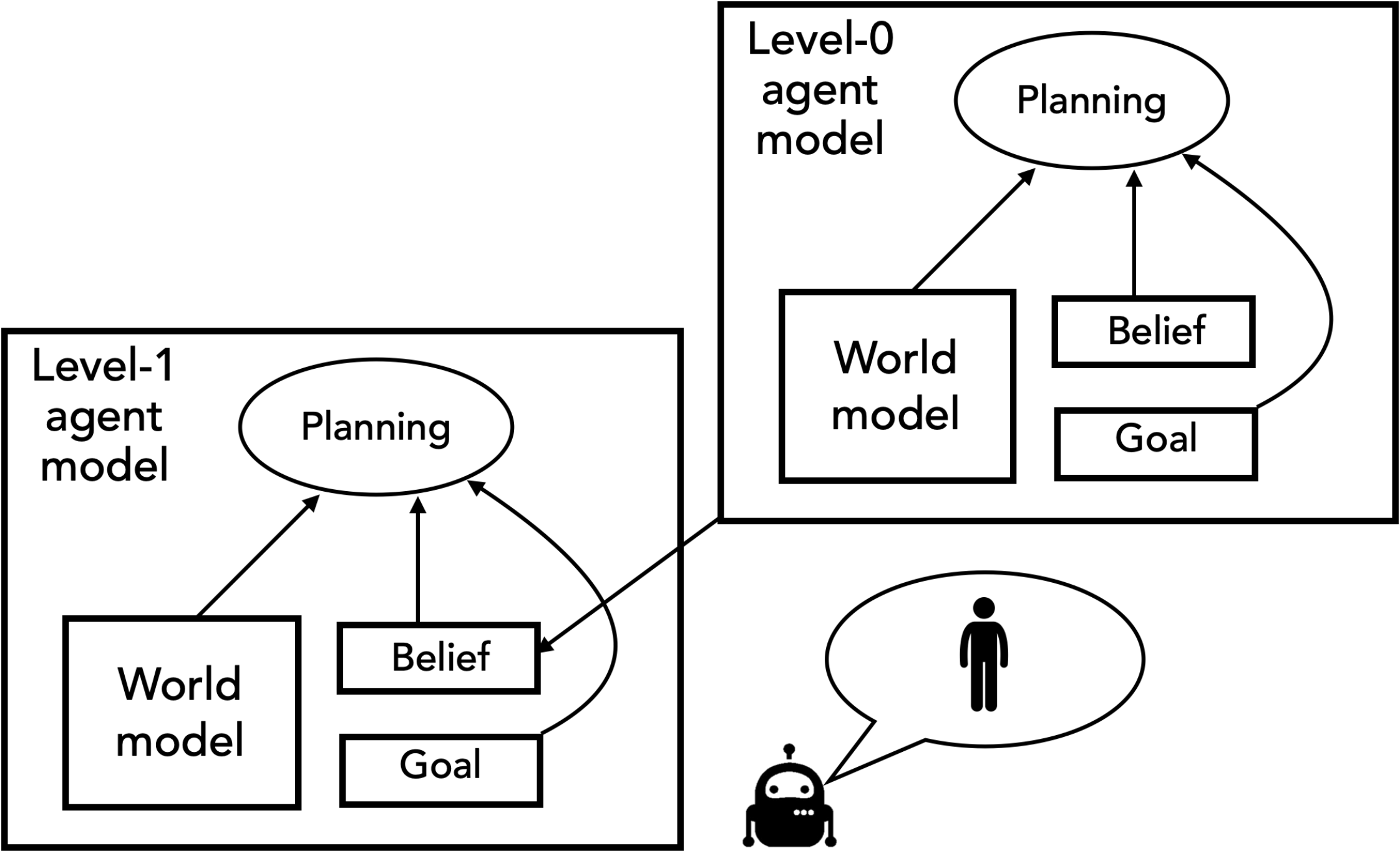}
    \caption{When an agent infers the mental state of another agent, it needs to build a mental model of another agent. This can be formulated as a level-1 agent model reasoning about a level-0 agent model.}
    \label{fig:social}
    \vspace{-1pt}
\end{figure}

We not only need to understand the world around us but also make intelligent decisions to achieve our goals by interacting with the world. Moreover, we also have to understand and interact with other agents. On the one hand, we understand the distinction between physical entities and an agent and have represented them in fundamentally different ways since infancy \cite{spelke2007core, gergely2003teleological}. On the other hand, we can also appreciate the fact an agent's behavior is contained by the world and that models of agents can not be separate from models of the world. A minimum definition of an agent model includes the following components:

{\bf Goal and reward.} An agent has its goal, which defines a reward function that guides the agent's goal-directed behavior. Sometimes, the reward function also includes the cost of the agent's actions.

{\bf Belief.} For an agent that has only partial observation of the world (e.g., a robot can only sense the objects around it), it has only incomplete information about the world state. Therefore, it needs to form a belief about what the true world state could be.

{\bf World model.} An agent has its world model in its mind, which may or may not be the same as the actual world. For instance, where we imagine a basketball will land after we throw it may be different from where it will actually land.

{\bf Planning.} Given an agent's mental state (goal, reward, and belief), its rational behavior can be modeled as planning which searches for actions that maximize its reward or reach its goal by simulating ahead using the world model in its mind.

There are two levels of use of agent models:

In embodied tasks, an agent model represents how an embodied agent optimizes its actions to maximize its accumulated reward based on its belief of the current world state and the physical constraints defined in its world model. For instance, given the command of ``give me a cup,'' a robot needs to find the cup as quickly as possible (goal and reward) based on where it believes the cup could be (belief) and the shortest path to reach the likely locations without hitting any obstacles (world model). We term this type of agent model {\it level-0} agent model. There have been works on using LMs to build level-0 agent models for language agents \citep[e.g.,][]{andreas2022language,sumers2023cognitive,deng2023mind2web} and embodied agents \citep[e.g.,][]{huang2022language, ahn2022can, li2022pre}. 

In social reasoning tasks, we utilize the models of \textit{other} agents to reason about their behaviors. This capacity is commonly referred to as \textbf{Theory of Mind} \citep{premack1978does}, which involves forming mental models of other agents and conducting causal reasoning to interpret other agents' behaviors in terms of their mental states (such as goals and beliefs). We term the agent models that reason about other agents, {\it level-1} agent models (Figure~\ref{fig:social}). For instance, to understand a person's searching behavior, we need to infer what goal (the object they are looking for) and belief (where they believe the object is) may lead to the plan (the observed behavior) of that person. Systems designed to interact with humans, such as assistive robots \citep[e.g.,][]{dautenhahn2007socially, hadfield2016cooperative,patel2022proactive,puig2023nopa}, AI teachers \citep[e.g.,][]{wang2021towards}, autonomous vehicles \citep[e.g.,][]{chandra2020stylepredict}, and cooperative embodied agents \citep[e.g.,][]{bara2021mindcraft,pmlr-v162-sclar22a},  must be able to understand and cooperate with humans in a grounded, physical world. Therefore, there is a critical need for AI systems to develop robust social reasoning that combines social commonsense (via level-1 agent models) and physical commonsense (via world models). Recent studies have revealed the lack of human-level social reasoning in LMs \citep[e.g.,][]{sap2022neural, jin2023mmtom, ullman2023large, shapira2023clever,moghaddam2023boosting}. We hypothesize that it is possible to enhance LMs' social reasoning capacity by building explicit world models and level-1 agent models. We may even enable recursive social reasoning \citep[e.g.,][]{gmytrasiewicz2005framework,goodman2016pragmatic,hadfield2016cooperative,tejwani2022social,schulz2023emergent,jha2023neural}) via higher-level agent models. 
\section{The LAW Framework}\label{sec:law}

\subsection{Reasoning with World and Agent Models, on the Language Model Backend}\label{sec:law:inference}

\subsubsection{Limitations of Reasoning with Language Models}

LLMs have exhibited strong reasoning abilities in many language tasks. Recent LM reasoning approaches further boost their performance by guiding LMs to generate intermediate reasoning steps. For example, Chain-of-Thought (CoT) \citep{wei2022chain}  prompts the LMs to generate step-by-step derivations before producing the final answer. More recent approaches introduce more sophisticated structures into the reasoning process, such as decomposing a target question into a series of subquestions \citep{zhou2022least,xie2023decomposition}, using beam or tree-structured search to find better reasoning chains \citep{Yao2023TreeOT,jung2022maieutic,zhu2022solving,liu2023llm+}, and adding self-verification steps for rectifying reasoning errors \citep{ouyang2023structured,shinn2023reflexion,madaan2023self,weng2022large}.

Compared to LLM reasoning, deliberate human reasoning relies on the internal world model which allows human brains to play out different reasoning steps and their effects on the world state. Take the example of playing BlocksWorld, which involves generating an action plan to rearrange blocks to a target configuration (Figure~\ref{fig:failures}, embodied reasoning). To devise such an action plan, humans imagine different potential actions (e.g., ``{\it pick up the red block}''), simulate the state (i.e., block configuration) after each action using the world model, and assess its likelihood of achieving the desired outcome. We then refine our action plan by choosing the most promising steps. Similarly, when solving a math problem, we explore different possible derivation steps and their resultant states (i.e., intermediate conclusions derived so far), evaluate how each state is closer to the final solution, and choose the best derivation path accordingly. In both cases, the internal world model plays a key role by allowing us to explore multiple possibilities, simulate their outcomes, and iteratively refine the reasoning trace.

Inspired by human reasoning, we can pinpoint several essential components that are missing in the current reasoning with LLMs, including:
(1) explicit modeling of the world state (e.g., block configuration, intermediate math conclusions); for example, as in Figure~\ref{fig:failures} (embodied reasoning), CoT typically generates a sequence of actions without describing the block configuration after each step, often leading to inconsistent action plans (such as those yielding invalid states); 
(2) an internal WM for simulating future states, which is a foundation of human reasoning; 
(3) a reward mechanism to assess and guide the reasoning towards the desired states; 
and (4) due to the above, balance between {\it exploration} (of possible reasoning options not considered yet) vs. {\it exploitation} (of the best reasoning steps identified so far), to efficiently navigate the vast reasoning space and find the optimal reasoning trace.

\subsubsection{Reasoning with World and Agent Models using Language Model Backend}

The above limitations call for a new conceptual framework of machine reasoning. Instead of reasoning directly with LMs, we propose that world and agent models are a better abstraction for carrying out robust and versatile reasoning. With the explicit, built-in components, including beliefs, anticipation of outcomes, and goals/rewards, a reasoning formulation based on world and agent models inherently overcomes the aforementioned limitations. Given a problem, the agent model performs deliberate planning in the reasoning space based on its beliefs about the current state and other agents as well as its prediction of future states resulting from various actions (through the world model), all directed by the agent's goal. The agent decides on the next step or an action plan by maximizing its reward while adhering to constraints due to its beliefs and world model. Under this abstraction, crucially, LLMs are used as the backbone for implementing the system or its components. Therefore, the system incorporates the computation power and flexibility of LLMs for processing the diverse noisy scenarios in the real world, and the structured abstraction of world and agent models to enable robust, efficient, and versatile reasoning capabilities in language, embodied, social, and other problems. In the remainder of this section, we review recent works that have made meaningful progress relevant to the proposed framework.
We discuss the limitations of the current LLM backend and outline the future research directions later.

\begin{figure}[t]
    \centering
    \includegraphics[width=0.85\linewidth]{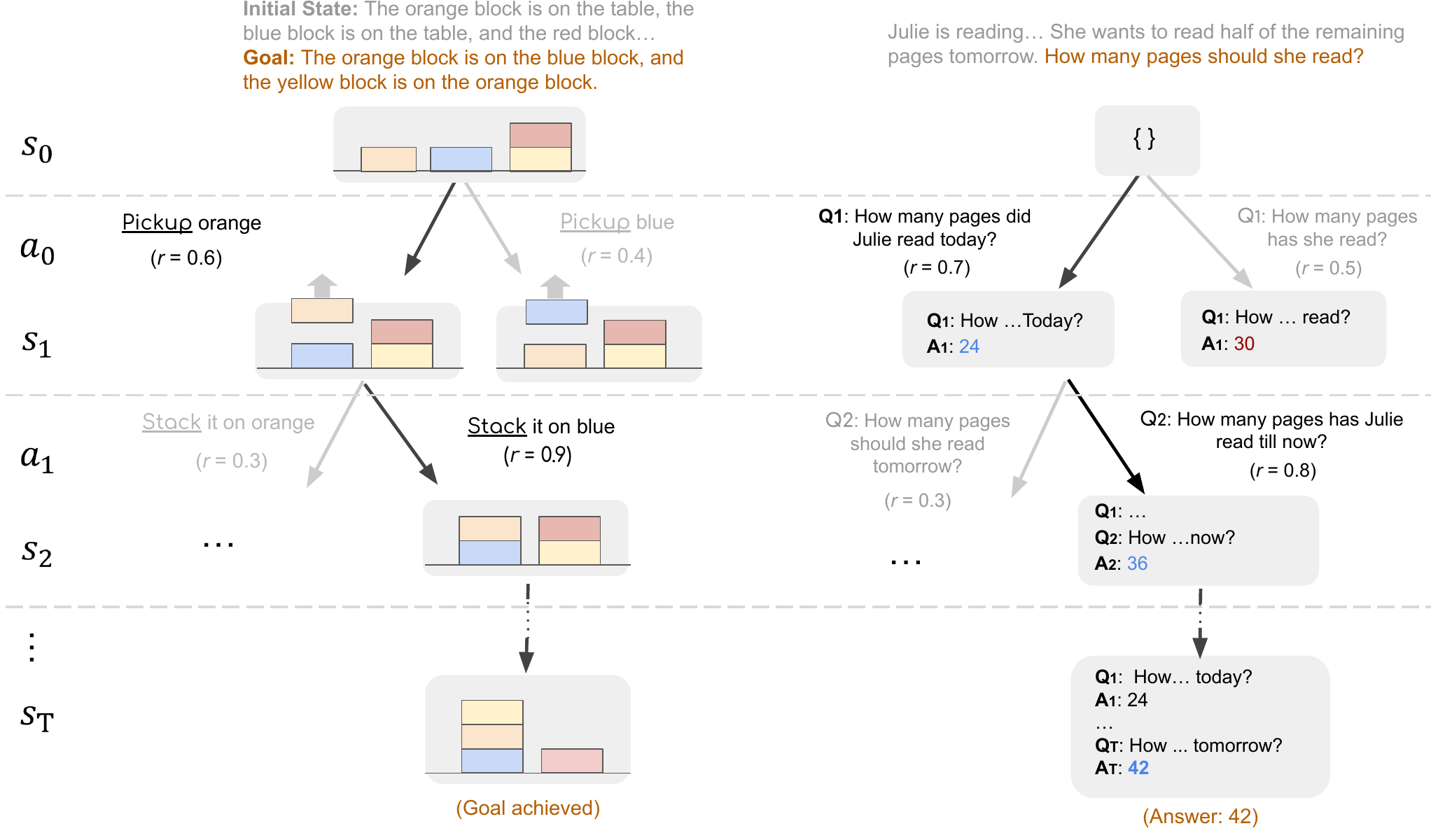}
    \caption{Illustration of RAP \citep{hao2023reasoning} for reasoning in BlocksWorld and math problems. (Figure from \citet{hao2023reasoning}).}
    \label{fig:rap}
    \vspace{-1pt}
\end{figure}

\paragraph{LMs as Both World and Agent Models.} 
Perhaps of most relevance to the \modelname framework is {\it Reasoning-via-Planning (RAP)} \citep{hao2023reasoning} which introduced the idea of world and agent modeling into the reasoning problems previously handled by LLMs directly (Figure~\ref{fig:rap}). Specifically, RAP repurposes an LLM as a world model by prompting the LLM to predict the next state $s_{t+1}$ of reasoning after applying a reasoning step $a_t$ to the current state $s_t$ (e.g., predicting new conclusions after a derivation step for a math problem, as described above). Similarly, the same LLM is prompted to act as an agent model that produces an action after each state. As a result, a reasoning trace consists of a sequence of interleaved states and reasoning steps $(s_0, a_0, s_1, \dots, a_{T-1}, s_T)$. This differs from the previous reasoning methods, such as CoT as mentioned above, where the reasoning focuses on generating a sequence of only actions, e.g., (\texttt{$a_0=$ ``pickup red block'', $a_1=$ ``stack on blue block'',} \dots). Similar as in \citep{li2022language}, augmenting the reasoning with the (predicted) world states helps the LM with a more grounded and coherent inference. The full reasoning trace is simulated by the LLM itself (as a reasoning agent with an {\it internal} world model) without interacting with the {\it external} real environment. This resembles humans contemplating a possible plan in their minds.

More crucially, the capability of simulating future states (due to the introduction of the world model) allows the incorporation of principled planning algorithms for strategic exploration in the vast reasoning space. RAP uses the classic Monte Carlo Tree Search (MCTS) \citep{kocsis2006bandit, coulom2007efficient} for finding high-reward reasoning traces with a balance between exploration and exploitation. Note that strategic search with MCTS was also used in previous successful systems such as AlphaGo \citep{silver2016mastering}. In problems like chess and Go, perfect world models exist (e.g., each move deterministically leads to a subsequent chess state). Real-world reasoning problems are more challenging due to the complex uncertain state dynamics. RAP and its follow-ups \citep[e.g.,][]{wang2023promptagent} show the benefits of structuring LLM reasoning with future state prediction and strategic planning.

As a general way to construct generative models, probabilistic programs have also been used for constructing world models and agent models for physical \citep[e.g.,][]{gothoskar20213dp3} and social reasoning \citep[e.g.,][]{zhi2020online}. A recent work \citep{wong2023from} leverages the code-writing capacity of LMs to translate natural language descriptions about the world and other agents to probabilistic programs of the world and other agents. This provides an alternative use of LMs in the constructing world and agent models, in which LMs serve as a flexible interface between language and thought (about the world and other agents).




\paragraph{LMs as the Planner in Agent Models.}
There have been many works in building embodied agents using LMs. The most common use of LMs is to generate plans based on prompts that specify the state, task, and even memory. While empirical results on LMs' planning capacity have been promising \citep{huang2022language}, the plans generated by LMs often fail to robustly solve long-horizon planning problems in complex, partially observable. To address this limit, recent works have been using LMs in an interactive planning paradigm, providing feedback from the environment and reflection on past actions as additional prompts for LMs to adjust their plan generation for future steps. Such an interactive planning paradigm has achieved success in both single-agent planning \citep[e.g.,][]{dasgupta2023collaborating, wang2023describe} and multiagent collaboration \citep[e.g.,][]{mandi2023roco, zhang2023building}. Finetuning LMs on specific domains has also been demonstrated to be beneficial for improving their planning capacities on the trained tasks. Specifically, the finetuned LMs exhibit a certain level of compositional generalization within the same domain \citep{li2022pre}. However, it remains unclear how much of the acquired planning capacity during finetuning can be generalized to novel domains. Moreover, when using the LMs for reasoning about the plans of \textit{other} agents (i.e., as the planners in other agents' models), we can see an improved Theory of Mind capacity compared to using LMs to directly infer other agents' mental states \citep{jin2023mmtom}. This suggests that while LMs on their own still lack social reasoning capacity, they can serve as a component in agent models to achieve better model-based social reasoning. Lastly, beyond embodied agents, LMs can also simulate social behaviors in abstract environments mimicking a simplified society \citep{park2023generative}. Without the need to generate physically grounded actions, LMs can synthesize high-level but also more sophisticated social behaviors. 

\paragraph{LMs as the Goal/Reward in Agent Models.} Although LMs have demonstrated promising planning abilities, for many embodied tasks (such as low-level robot control), conventional methods still have better performance. Instead of using LMs to produce the final plans, recent works have studied the possibility of using LMs as a component in an agent model, most notable for generating goals \citep{xie2023translating} or rewards \citep{yu2023language,kwon2023reward,ma2023eureka}. Goal and reward specifications grounded to a physical robot body for intended tasks can be difficult and typically require expert knowledge. However, the in-context learning capacity of LMs can provide an easier way to translate language descriptions about the intended tasks to accurate goal and reward specifications following a few provided examples.

\paragraph{LMs as the Belief in Agent Models.} To the best of our knowledge, there has not been much work on explicating modeling beliefs using LMs as a separate module. However, there have been evaluations of LMs' ability to encode belief representations about the world states \citep[e.g.,][]{li2021implicit}, showing promising but imperfect results. Additionally, there has been empirical evidence showing LMs' lack of ability to infer other agents' beliefs using Theory of Mind benchmarks \citep{sap2022neural,jin2023mmtom,ullman2023large,shapira2023clever}. For future work, it could be valuable to explicitly model belief update for an agent model as a separate module using LMs, similar to LMs as the planner, goal, or reward.




\subsection{Enhancing the Language Model Backend}\label{sec:law:learning}

The new perspective of reasoning under the \modelname framework also reveals a number of directions for enhancing the LM backend, in order to better operationalize the reasoning system or its modules. In particular, LMs need to learn by not only imitating existing data corpora but also all different forms of experience \citep{Hu2022Toward}, such as interacting with external environments and other agents, to gain a more robust and comprehensive understanding of the physical and social world. Moreover, as discussed previously, language is often not the most efficient medium for expressing all information during reasoning (e.g., describing a world state in world modeling). This calls for multi-modal understanding and generation capabilities in the backend model, to support more versatile and grounded world and agent modeling during reasoning. As we discuss below, recent studies have begun exploring these areas, yet there is still considerable room for further advancements.

\paragraph{Learning with Embodied Experiences.} Learning from pure text is unlikely to be sufficient to acquire much of the knowledge of the physical world and develop robust embodied skills. Recent works have explored the possibility of enhancing LMs' world knowledge with embodied experiences. Recent works have proposed different ways to collect embodied experiences, including random exploration \citep{xiang2023language}, accomplishing specified goals \citep{xiang2023language,zeng2023agenttuning,wang2023mint}, and proposing new tasks for an LM itself via an auto-curriculum \citep{wang2023voyager}. These diverse embodied experiences can unlock new ways to train language models to acquire knowledge about the world, with objectives beyond instruction finetuning and simple human preference feedback \citep[e.g., RLHF, ][]{ouyang2022training}.

Given collected embodied experiences, we can finetune LMs for domain-specific tasks \citep[e.g.,][]{li2022pre} and only use the resulting models for the target domains. However, it is also possible to preserve LMs' original language skills while injecting the additional embodied knowledge into the LMs, as studied in \citep{xiang2023language}. Finally, instead of finetuning LMs, \cite{wang2023voyager} have also explored the possibility of constructing an ever-growing repository of skills through memory.

\paragraph{Learning with Social Interactions.} In addition to learning from embodied experiences, we hypothesize that LMs can also benefit from social learning. For instance, LMs can learn from (1) observing human demonstrations for performing embodied tasks, (2) observing human social interactions, and (3) interacting with humans or other models (including LMs, \citet{liu2023training}). Such social learning experiences would not only help LMs acquire world knowledge from humans and other LMs but also develop better agent models that can support stronger social reasoning.

\paragraph{Multimodal World Modeling.} 
As discussed earlier, language has only a limited capacity to describe the world state and its dynamics. Therefore, there is a need for multimodal processing for world models (and agent models). One way to achieve this is to learn multimodal models, such as GPT-4V, LLaVA \citep{liu2023visual}, and Gemini \citep{Gemini2023}. While these models could be powerful tools for many tasks (especially for multimodal understanding), they are limited to act as world models due to the inability of {\it generating} images/videos for describing world states sequentially.

Recent advances in generative models such as diffusion models have provided a new way of modeling the world -- learning a video generator that can predict the future frames conditioned on action commands \citep[e.g.,][]{yang2023learning,hu2023gaia}. Such video prediction-based world models can simulate the detailed change in the world state, allowing motion planning that is unable to be achieved by world models constructed by LMs alone. However, training long-horizon video prediction models that can generalize to novel scenarios is difficult. It can also not be efficient, 
as the frame-level simulation is only necessary for low-level motion control, whereas, for high-level task planning, abstract state representations are adequately sufficient.
Therefore, we can envision a multi-level multimodel world model,
simulating the world at an abstract level (e.g., symbolic state representations such as scene graphs) and fine-grained level (e.g., pixels or other types of raw sensory data).

\paragraph{Tool Using. }
Enabling LMs to use external tools (e.g., functions, APIs, other models) serves as another way to augment LMs with multimodal capabilities \citep{AutoGPT,openai2022chatgptplugins}. Emerging research has been done on building LM agents that use tools for completing various tasks, through finetuning LMs with tool-use demonstration data \citep{schick2023toolformer,patil2023gorilla}, in-context learning \citep{yao2023react,paranjape2023art}, tool embedding \citep{hao2023toolkengpt}, and others. Most works still rely on LMs to perform direct reasoning and determine the application of tools within the process. We expect the world/agent model abstraction will facilitate enhanced tool-using capabilities.




\section{Discussions}

We presented the \modelname framework as a new perspective of formulating machine reasoning. Integrating the crucial elements of belief, future anticipation, goals/reward, and strategic planning, \modelname aims at more robust and versatile reasoning capabilities beyond the current reasoning with language models.
Aspects of the \modelname framework are aligned with recent proposals about building world models \citep{lecun2022path} and agent models \citep{andreas2022language}. Crucially, \modelname introduces an integrated framework that combines three models in a cognitively grounded way for solving a broad range of tasks. We have discussed how existing language models may serve as the backend for reasoning with world and agent worlds. We have also proposed possible ways to enhance the world and agent modeling capacity of the language model backend, including new training paradigms and the augmentation of multimodality capabilities.

We recognize that the \modelname framework has its limitations. First, the language model backend implies symbolic representations in a discrete space. We have discussed the possibility of augmenting this space with additional continuous latent spaces modeled by other modalities (e.g., the latent space for a diffusion model that simulates pixel-level world state transitions). However, it may also be possible to use a single continuous latent space for a world model or an agent model. While we hypothesize that symbolic representations from language models may help us to learn the causal structures of the world and agents as demonstrated by existing LMs, it remains unclear whether continuous latent representations can achieve the same capacity \citep{ha2018world, hafner2019learning, anand2019unsupervised, ermolov2020latent, lecun2022path}. Second, it is possible that the current world and agent modeling may not capture all knowledge about the world and agents. For instance, we assume that agent behaviors are driven by goals or rewards. However, behaviors can be driven by other variables, such as social norms. Lastly, this paper does not discuss the inherent limits of Transformer architectures \citep[e.g.,][]{dziri2023faith}. We believe that further studies on understanding the learning mechanism of Transformers can be complementary to and beneficial for the development of machine reasoning.


\clearpage
\bibliography{neurips_2023.bib}
\bibliographystyle{abbrvnat}

\end{document}